\documentclass{article}
 \pdfoutput=1
    \PassOptionsToPackage{sort, compress, numbers}{natbib}

    \usepackage[preprint]{neurips_2019}

\usepackage[utf8]{inputenc} 
\usepackage[T1]{fontenc}    
\usepackage{hyperref}       
\usepackage{url}            
\usepackage{booktabs}       
\usepackage{amsfonts}       
\usepackage{nicefrac}       
\usepackage{microtype}      

\usepackage{amsmath,amsfonts,amssymb,amsthm}
\usepackage{caption, rotating, graphicx}
\usepackage{subcaption}
\usepackage{xcolor}
\usepackage[subpreambles=true]{standalone}

\newcommand{\norm}[1]{\left\lVert#1\right\rVert}
\newcommand{\setnotation}[1]{\mathcal{#1}}
\newcommand{\dataindex}[3]{{#1}_{#2}^{(#3)}}
\newcommand{\parametric}[3]{#1(#2; #3)}
\newcommand{\shortparam}[2]{#1(#2)}
\newcommand{\risk}[2]{\mathbb{E}_{#1}\left[#2\right]}
\newcommand{\seq}[2]{\{#1\}_{#2}}

\newcommand{\smoothindfunc}[1]{\left(#1\right)_{+}}
\newcommand{\loss}[2]{\ell(#1, #2)}
\newcommand{\reg}[2]{R_{#1}(#2)}

\newcommand{\U}{\mathbf{U}}

\newcommand{\ourtitle}{Implicit Label Augmentation on Partially Annotated Clips via Temporally-Adaptive Features Learning}

\title{\ourtitle}

\author{%
   Yongxi Lu \qquad Ziyao Tang \qquad Tara Javidi \\
   University of California, San Diego \\
   \texttt{\{yol070, zit021, tjavidi\}@eng.ucsd.edu}
}

\begin{document}

\maketitle

\begin{abstract}

  Partially annotated clips contain rich temporal contexts that can complement the sparse key frame annotations in providing supervision for model training. We present a novel paradigm called Temporally-Adaptive Features (TAF) learning that can utilize such data to learn better single frame models. By imposing distinct temporal change rate constraints on different factors in the model, TAF enables learning from unlabeled frames using context to enhance model accuracy. TAF generalizes ``slow feature'' learning and we present much stronger empirical evidence than prior works, showing convincing gains for the challenging semantic segmentation task over a variety of architecture designs and on two popular datasets. TAF can be interpreted as an implicit label augmentation method but is a more principled formulation compared to existing explicit augmentation techniques. Our work thus connects two promising methods that utilize partially annotated clips for single frame model training and can inspire future explorations in this direction. 

\end{abstract}

\section{Introduction}

The success of modern machine learning techniques in solving challenging problems such as image recognition depends on the availability of large-scale, well-annotated datasets. Unfortunately, the most complex and useful tasks (e.g. semantic segmentation) are usually also the ones that require the most labeling efforts. This is arguably a major obstacle for large-scale applications to real-world scenarios, such as autonomous driving, where model performance is critical due to safety concerns. In this work, we focus on methods that can utilize partially annotated clip data, more precisely short video sequences with annotations only at key frames, to improve model performance. Datasets in this format are natural byproducts of typical data collection procedures. From clips, a large number of unlabeled frames is available at virtually no additional cost. But clip data can nevertheless encode rich temporal contexts useful for training more accurate models. Fully utilizing partially annotated clips in learning is an interesting problem not only for its practical relevance, but also because it provides partial answers to an interesting scientific question: Humans can naturally learn from continuous evolution of sensing signals without much ``labels'', can machines do the same? 


We investigate a particularly intriguing case: To train a model that benefits from temporal information during training but is used to make predictions on independent frames at inference. This is in contrast to video prediction models \cite{Jampani2017VideoPN,8296851,10.1007/978-3-319-54407-6_33,Nilsson2018SemanticVS,Gadde2017SemanticVC,Wang_2015_ICCV,Srivastava:2015:ULV:3045118.3045209,8237857,Mathieu2016DeepMV,10.1007/978-3-319-46478-7_51} where video clips are used at both training and inference. The main intuition of our approach is to decouple fast-changing factors and slow-changing factors in data.  Fast-changing factors reflect rapid temporal dynamics and can only be learned from a labeled frame or its immediate neighbors, while slow-changing factors can be learned from data points within a larger temporal context. Our method utilizes the temporal context provided by the partially annotated clips to learn better features without diminishing the ability to learn fine-grained features with rapid temporal changes. This is achieved by allowing different parts of the model to adapt to distinct temporal change rates in data, a.k.a. \textbf{Temporally Adaptive Features (TAF)} learning. We propose a principled approach to formalize this intuition by introducing temporal change rate constraints in the learning problem and show that the resultant optimization problem can be efficiently approximated by a feature swapping procedure with contrastive loss. The TAF paradigm generalizes the well-motivated ``slow feature'' learning methods \cite{7410822,Wiskott:2002:SFA:638940.638941} for self-supervised learning. In this regard, ours is the first to demonstrate significant empirical gains on a challenging real-world application via imposing temporal coherence regularization. It can also be seen as a form of implicit label augmentation and is related to explicit pseudo label generation techniques \cite{Mustikovela2016CanGT,8265246,Zhu2018ImprovingSS,8206371} which also show promising improvements in practice. But ours is a more principled treatment that handles the important issue of label uncertainty automatically. Interestingly, our work is the first to combine these two seemingly unrelated line of research. It thus sheds new light on the theory and practice of the important problem of learning from partially annotated clips and can benefit future explorations on this topic.

The TAF framework can in theory be applied to any recognition tasks with partially annotated clip data. However, the advantage in doing so will well depend on the task. We identify \textbf{semantic segmentation}, the task of assigning class labels to every pixel in an image, as a good test case due to the necessity of multi-scale modeling. Natural images usually feature structures with a great variety of sizes, functions and perspectives. This results in different intrinsic spatial and temporal change rates of different structures. A useful semantic segmentation model needs to provide comprehensive understanding of all these different structures. TAF can address this challenge by allowing different parts of the model to learn features with varying temporal change rates, rather than forcing all the features to vary slowly, as is the case of slow feature learning \cite{7410822,Wiskott:2002:SFA:638940.638941}. Beyond this particular task, semantic segmentation is also a good example of the broader set of ``dense prediction tasks'' in computer vision, such as object detection \cite{Alpher19, liu2016ssd, 7780627, 8099589, 8237586, Law_2018_ECCV, 8237584}, pose estimation \cite{6909610}, monocular depth estimation \cite{8100183,Zou_2018_ECCV,Yin2018GeoNetUL,Godard2017UnsupervisedMD,Garg2016UnsupervisedCF,Fu2018DeepOR}, instance segmentation \cite{8237584, Alpher19c, Alpher19d, Alpher19e} as well as panoptic segmentation \cite{DBLP:journals/corr/abs-1801-00868, DBLP:journals/corr/abs-1901-03784}, to name a few. Dense prediction tasks all share the key properties of laborious annotation and multi-scale features thus it is likely that our finding from semantic segmentation can directly benefit these tasks. 

This paper is organized as follows. Section \ref{sec:method} presents our method. Section \ref{sec:related} compares our method to related works. Section \ref{sec:exp} presents our empirical findings and ablation studies. Section \ref{sec:conclusion} concludes the paper and discusses future directions. 

\section{Methods}
\label{sec:method}

We first introduce notations useful to our presentation. We denote the dataset as $\setnotation{D} = \{x_i, y_i\}_{i=1}^{N}$. Each input and its associated labels can be finely indexed as $\dataindex{d}{t}{k}=(\dataindex{x}{t}{k}, \dataindex{y}{t}{k})$, where $k, t$ denotes the clip index and the time index within the clip, respectively. Whenever it is clear from the context, we use $(x_t, y_t)$ to denote input-label tuples at time $t$ for any particular clip.


\subsection{Temporally Adaptive Feature Learning}


Our method decouples the fast and slow changing factors in data by forcing the model to learn features that are adaptive to the varying temporal change rates. To be applicable to our framework, we assume the labeling function $y$ can be factorized as $\parametric{y}{x}{\Theta} = \parametric{\Omega}{\parametric{\Phi_1}{x}{\theta_1}, \cdots, \parametric{\Phi_m}{x}{\theta_m}}{\omega}$. We can quantify how fast the labeling function changes w.r.t. time by taking its time derivative. 

\begin{equation}
    \norm{\frac{d \parametric{y}{x_t}{\Theta}}{dt}} =
    \norm{\sum_{i=1}^m \frac{\partial \Omega}{\partial \Phi_i(x_t)}\frac{d \Phi_i(x_t)}{dt}}
 \triangleq \norm{\sum_{i=1}^m \Psi_i(x_t)}
\end{equation}

Note that $y$ can be seen as a function with $m$-dimensional input where $\parametric{\Phi_i}{x_t}{\theta_i}$ represents one of its dimensions. $\Psi_i(x_t)$ quantifies the variation of $y$ w.r.t. time through this dimension. The ``fast'' and ``slow'' factors are characterized by the degree at which they contribute to temporal variations in the predictive model $y$. To instantiate this idea, our TAF frameworks solves the following empirical risk minimization problem with temporal change rates constraints. 

\begin{subequations}
\begin{alignat}{2}
&\underset{\Theta}{\min}    &\qquad& \risk{\setnotation{D}}{\loss{\parametric{y}{x}{\Theta}}{y}} \\
&\text{subject to} &      & \norm{\Psi_i(x_t)} \leq c_i, \quad \text{where } 0 \leq c_1 \leq c_2 \leq \cdots \leq c_m.
\end{alignat}
\end{subequations}

For a differentiable model, the analytical form of the constraints is available if $\frac{d x_t}{d t}$ is provided. In applications where $x_t$ is a high dimensional vector (such as an image), this may not be possible. Thus, we propose to use first-order finite difference to approximate the constraints via neighboring samples. 

\begin{equation}
    \norm{\Psi_i(x_t)} \approx
    \frac{
    \norm{
    \shortparam{\Omega}{\seq{\shortparam{\Phi_j}{x_{t}}}{j \neq i}},\shortparam{ \Phi_i}{x_{t+\Delta}} - 
    \shortparam{\Omega}{\seq{\shortparam{\Phi_j}{x_{t}}}{j \neq i}},\shortparam{ \Phi_i}{x_{t}}}
    }{|\Delta|}
    \triangleq \frac{\delta y(x_t, i, \Delta)}{|\Delta|}
\end{equation}

The constrained optimization problem itself is difficult to solve. We can convert it into an unconstrained optimization with the following regularization term which approximate the original constraints. This permits the use of gradient-based solvers if the model is differentiable. 

\begin{equation}
\label{eqn:objective}
    \reg{+}{x_t, i, \Delta} = \smoothindfunc{\delta y(x_t, i, \Delta) - |\Delta|c_i} = \max \left(0, \delta y(x_t, i, \Delta) - |\Delta|c_i \right)
\end{equation}

The proposed regularization is defined on any pairs of samples separated by known interval $\Delta$, even if their labels are unknown. This construction thus enables learning from unlabeled data. Needless to say, TAF learning is limited to short clips as the first order approximation is valid only for small $\Delta$. The slack term $|\Delta|c_i$ promotes features that adapts to a distinct temporal change rate. When $c_i$ is small, that dimension is forced to model slow-changing factors shared within a large temporal context, which is an implicit form of data augmentation. The dimensions with large $c_i$ on the other hand can still model rapid motions in data important for the task. As we will discuss in ablation studies and supplementary materials, $c_i$ and $m$ are important hyper-parameters.  

As a remark on related methods, we note that Eqns. \ref{eqn:objective} generalizes the temporal coherence regularization in \cite{7410822} to multiple change rates, making it more suitable for real-world applications such as semantic segmentation where multi-scale features are essential. This regularization can also be seen as implicitly assuming constant labels across the entire clip, with the slack term acknowledging the uncertainty introduced by this approximation. In this regard, $c_i$ measures the growth rate of uncertainty in time of the implicit pseudo labels. Prior works suggests that properly modeling the uncertainty in pseudo labels to be important in the final task performance \cite{Mustikovela2016CanGT}. While TAF learning is motivated differently, it leads to a similar construction.

Finally, $\setnotation{D}_{\text{key}} \subseteq \setnotation{D}$ denotes the subset of the data with annotations, the half-length of each clip is $n_h$, $\Delta_0$ is the sampling period and $\U$ denotes the uniform distribution defined on integers. The regularized optimization problem becomes

\begin{equation}
\label{eqn:raw_optimization}
    \underset{\Theta}{\min} \quad \risk{(x, y) \sim \setnotation{D}_{\text{key}}}{\loss{\parametric{y}{x}{\Theta}}{y}} + 
    \lambda \risk{\substack{(x_{t}, -) \sim \setnotation{D}, \\ n \sim \U(-t \Delta_0, 2 n_h - t \Delta_0)}}{\sum_{i=1}^m \left[\reg{+}{x_t, i, n \Delta_0} \right]}.
\end{equation}

\begin{figure*}
    \centering
    \includegraphics[width=0.95\linewidth]{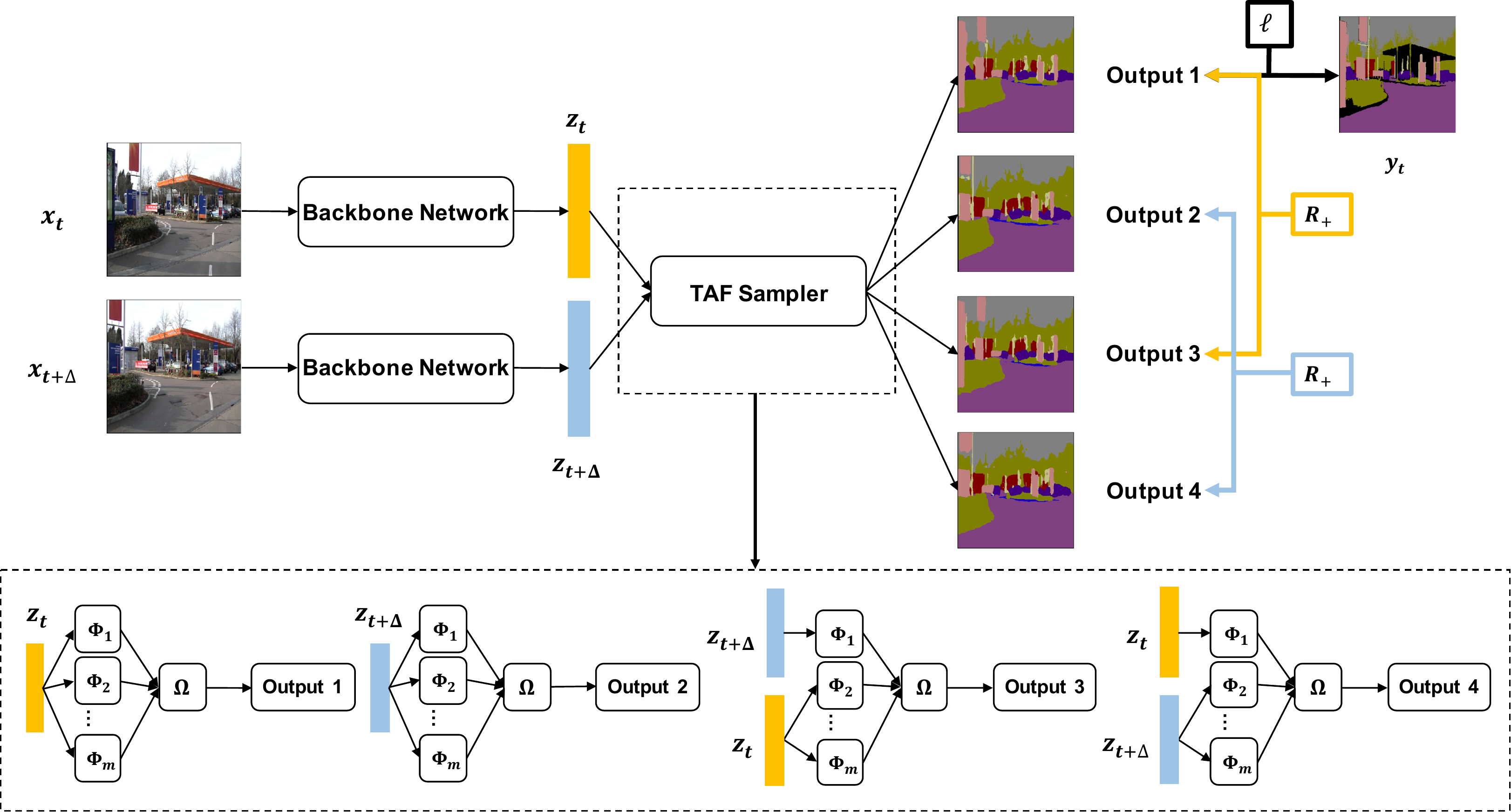}
    \caption{The sampling procedure of TAF learning for a single pair of images, in our implementation for semantic segmentation where $\Phi_i$ are defined on features extracted from a shared backbone. }
    \label{fig:model}
\end{figure*} 

\subsection{Efficient Frame Sampling}

The proposed objective function in Eqns. \ref{eqn:raw_optimization} is computationally inefficient when combined with mini-batch SGD or its variants. First of all, the computation of the sample averages requires two separate sampling streams: One for the key frames with annotations for the loss function, and the other for the regularization term using pairs of frames. In general, there is no ensured overlapping in these two streams of samples. As a result, we usually cannot use features computed from an image to update both  terms. This inefficiency is exacerbated by the fact that the regularizer requires pair inputs, making the training even less efficient. Secondly, the regularization term requires separate feature exchanges for each feature dimension. When $m$ is large and the model decoupling requires re-computation of a significant portion of the model, this strategy is highly inefficient. 

Our proposal is as follows: Within each mini-batch, a set of image-label tuples are first sampled from the annotated key frame subset $\setnotation{D}_{\text{key}}$. Each of these tuples are associated with a clip. Then, for each key frame sampled a random (unlabeled) pairing image is selected from the same clip by sampling the index difference between the key frame and the unlabeled pairing frame. In this improved procedure, all feature computations contribute to all terms in the objective function. To further make use of cached features, the regularization term is also made symmetric. To ensure tractable mini-batch updates, the summation over the dimensions are replaced by a uniform sampling of the dimension index $i$ at each training example. The efficient TAF procedure solves the following problem

\begin{equation}
    \underset{\Theta}{\min} \quad \risk{\substack{(x_t, y_t) \sim \setnotation{D}_{\text{key}}, \\ n \sim \U(-n_h, n_h), \\ i_1, i_2 \sim \U(1, m)}}{\loss{\parametric{y}{x_t}{\Theta}}{y_t} + \lambda \left[\reg{+}{x_t, i_1, n \Delta_0} + \reg{+}{x_{t+n\Delta_0}, i_2, -n \Delta_0} \right] }
\end{equation}

where we simplify the notation by assuming that the key frame is always at the center of each clip. Figure \ref{fig:model} illustrates how the training objective is computed between a pair of sampled images.

The number of training iterations of TAF learning is two times of the baseline as only half of the mini-batch have ground truth labels. In order to compute the pair-wise loss, the aggregation function $\Omega$ has to be evaluated twice in each forward pass \footnote{The first time using the original features, the second time using features after swapping.}. Thus for tractable training, $\Omega$ should be chosen to be a lightweight function. Figure \ref{fig:model} illustrates the proposed sampling procedure in the application of semantic segmentation, where we use a Siamese network for the backbone feature extractor and apply the TAF procedure only at the encoder layers (details in Section \ref{sec:app_semantic}). 

\subsection{Application to Semantic Segmentation}
\label{sec:app_semantic}

We now provide a brief overview of two of the most popular semantic segmentation models and explain how our framework can be applied. The multi-branch structure that enable TAF learning for FCNs and DeepLab v3+ is used in a broader set of architectures for semantic segmentation \cite{8100143,8099589,Yu2016MultiScaleCA,Chen2017RethinkingAC} and we expect similar modifications to be feasible. 

\paragraph{FCNs} Fully convolutional networks (FCNs) \cite{7298965} is one of the earliest and most popular deep-learning based architecture for semantic segmentation. It follows a straightforward multi-scale design: Feature maps at the output of three different stages of a backbone convolutional network are extracted. Due to the downsampling operators between stages, feature maps have a decreasing spatial resolution (in the case of FCN8s that we consider the output stride equals to 8, 16 and 32, respectively). The features maps are converted into class logits maps via a single layer of convolutions. The three predictions are aggregated via a cascade of upsampling and addition operations. In our modification of FCNs we assign $\Phi_1(x), \Phi_2(x)$ and $\Phi_3(x)$ to represent three feature maps, where $\Phi_1(x), \Phi_2(x), \Phi_3(x)$ represents the stride-32, stride-16 and stride-8 feature maps respectively. The aggregation function is the single convolution layer and the following cascaded addition operations. In practice, we find swapping $\Phi_1(x)$ is sufficient for improved accuracies over the baselines. 

\paragraph{DeepLab v3+} DeepLab v3+ \cite{Chen2018EncoderDecoderWA} is a recent semantic segmentation algorithm that has achieved state-of-the-art accuracy in challenging datasets such as Pascal VOC and Cityscapes. It follows an encoder-decoder structure, where the encoder is an ASPP module \cite{7913730} that consists of five branches with different receptive fields (modeling structures at different scales): Four branches with varying dilation rates and an additional image pooling branch. Similar in spirit to the FCNs case, we assign $\Phi_1$ to the image pooling branch, and $\Phi_2 \cdots, \Phi_5$ to the remaining branches starting from the one with largest dilation rate. The decoder is the aggregation function $\Omega(\cdot)$ in our formulation. 

\section{Related Works}
\label{sec:related}

\paragraph{Regularization and Data Augmentation in Deep Neural Networks} There is a rich literature of generic regularization and data augmentation techniques sharing our goal of improving generalization, e.g. norm regularization \cite{NIPS1991_563, Ng:2004:FSL:1015330.1015435}, reduction of co-adaptation \cite{JMLR:v15:srivastava14a,Wan:2013:RNN:3042817.3043055,devries2017cutout} and pooling \cite{6144164,pmlr-v51-lee16a,8099909,8099909}. For semantic segmentation, data augmentations techniques based on simple image transformations \footnote{such as horizontal flipping, random cropping, random jittering, random scaling and rotation} are standard practices. Recently, \cite{DBLP:journals/corr/abs-1805-09501,Hauberg2016DreamingMD, DBLP:journals/corr/abs-1902-09383,NIPS2017_6916} learn optimal transformations. These techniques are limited by not using video information but are complementary to our approach. We follow the default choice of regularization and random transformations when comparing TAF learning with corresponding baselines. Another effective solution is to use generative models for data and label synthesis \cite{8099724,8363576,Antoniou2018DataAG,DBLP:journals/corr/abs-1810-10863}. Similar to ours, these methods can improve model accuracy using unlabeled data. However, it could be intrinsically difficult to generate realistic and diverse data for complicated applications, while our method can directly utilize the large amount of real video clips. 

\paragraph{Single Image and Video Semantic Segmentation} We use semantic segmentation \cite{Chen2018EncoderDecoderWA,7913730,7298965,8100143,Yu2016MultiScaleCA,Yu2017DilatedRN,Chen2017RethinkingAC} as an example to verify our method as discussed in Section \ref{sec:related}. Importantly, our method is quite different from the related literature of video semantic segmentation \cite{Jampani2017VideoPN,8296851,10.1007/978-3-319-54407-6_33,Nilsson2018SemanticVS,Gadde2017SemanticVC,Wang_2015_ICCV,Srivastava:2015:ULV:3045118.3045209,8237857,Mathieu2016DeepMV,10.1007/978-3-319-46478-7_51}, where video clips are utilized at both training and inference. Our work learns models using video clips at training, but the model can be used on independent frames at inference. In semantic segmentation, unlabeled frames can be used via future frame predictions \cite{Srivastava:2015:ULV:3045118.3045209,8237857,Mathieu2016DeepMV,10.1007/978-3-319-46478-7_51,Vondrick2015AnticipatingTF} and label propagation \cite{Mustikovela2016CanGT,8265246}. The former is only shown to improve video prediction results but not on single image predictions (as expected as future frame prediction is difficult from a single frame due to the lack of temporal context at test time). A few preliminary works suggest the latter can bring promising improvements to single frame predictions by generating pseudo labels \cite{Mustikovela2016CanGT,8265246,Zhu2018ImprovingSS,8206371}. But video propagation notably relies on manual screening and careful hyper-parameter tuning to reject low quality labels \cite{Mustikovela2016CanGT}, otherwise it could surprisingly lead to performance degradation after including pseudo-labels in some cases \cite{8265246}. Our method has the advantage of not requiring manual intervention. More importantly, our work suggests that regularizing the temporal behavior of features is an implicit form of video augmentation without explicit modeling of temporal dynamics, which compared to video propagation is a simpler pipeline and could be more transferable to other tasks.

\paragraph{Self-Supervised Learning} Self-supervised learning utilizes the large amount of unlabeled data via carefully designed ``pretext'' tasks or constraints that aim at capturing meaningful real-world invariance structures in the data. The goal is to learn more robust features. Future frame prediction \cite{Mathieu2016DeepMV,10.1007/978-3-319-46478-7_51,Vondrick2015AnticipatingTF,8237857,Srivastava:2015:ULV:3045118.3045209}, patch consistency via tracking \cite{Wang_2015_ICCV}, transitive invariance \cite{xiaolong_iccv_17}, temporal order verification \cite{Misra2016ShuffleAL} and motion consistency \cite{Jayaraman2017, 7780548, 8100121} have been proposed as useful pretext tasks. Recently, consistency across tasks are also explored \cite{Ren2018CrossDomainSM, Doersch2017MultitaskSV}, although these methods do not consider videos. However, as pretext tasks usually differ from the target task, a separate transfer learning step is required. Ours in contrast can be used directly on the target task. Via imposing geometric constraints, several recent works use self-supervised learning to directly address real-world tasks, most notably in depth and motion predictions \cite{Garg2016UnsupervisedCF,8100183,Mahjourian_2018_CVPR,Mahjourian_2018_CVPR,DBLP:journals/corr/abs-1812-05642,Godard2017UnsupervisedMD,Jiang_2018_ECCV,Zou_2018_ECCV}. However, these methods cannot transfer easily outside of their intended geometry application. Among them, SIGNet \cite{DBLP:journals/corr/abs-1812-05642} points to a unified framework for self-supervised learning of both semantic and geometric tasks which would broaden the applications of this line of works, but the existing work can only improve on geometric tasks. In contrast, TAF is not restricted to any particular task by design. Our TAF framework generalizes the temporal coherence regularization in \cite{7410822,Wiskott:2002:SFA:638940.638941} to multiple change rates and is the first to validate the utility of this form of regularization on challenging real-world applications. In contrast, the prior works focus on theoretical insights and are not rigorously validated.

\section{Experiments}
\label{sec:exp}
\subsection{Datasets and Evaluation Metrics}
We test our approach on two widely-used datasets for semantic segmentation: Camvid \cite{BrostowSFC:ECCV08} and Cityscapes \cite{Cordts2016Cityscapes}. The images of both datasets are frames captured from videos. Detailed annotations are provided on key frames. The meta-data of the datasets include the source frame ids of the annotated frames which makes unlabeled frames within the same clips available. The availability of unlabeled frames in the said clip format makes these two datasets ideal for testing our TAF framework. In particular, Camvid consists of 367 clips for training and 101/233 images for val/test. Key frames from training and test set are captured at 1Hz and annotated with 11 object classes. We capture extra frames around the key frames at 30Hz using the provided raw video. Cityscapes consists of 2975 training key frames and 500 validation images. The key frames are the 20-th frames in the provided 30-frame clips (30Hz) annotated with 19 object classes. In the interest of fast experimentation and to test our methods on small datasets, we sample 20\% and 50\% of the clips from Cityscapes training set, creating customary datasets with 595 and 1488 training clips respectively. We follow standard evaluation protocols and report mIOU and pixel accuracy on the held-out set. 

\subsection{Comparison to Baselines}

\begin{table*}[t]
\begin{center}
\small 
\setlength\tabcolsep{3.5pt}
\renewcommand{\arraystretch}{1.25}
\begin{tabular}{lcccccc}
\hline \hline
\textbf{Method} & \textbf{Output stride} & \textbf{Training set} & \textbf{Backbone} & \textbf{TAF} & \textbf{mIOU ($\%$)} & \textbf{Pixel acc. ($\%$)}\\ 
\hline \hline
\multicolumn{7}{c}{Camvid test set}\\
\hline
FCN8s & 8,16,32 & Camvid & ResNet-50 & & $65.3$ & $91.0$ \\
FCN8s & 8,16,32 & Camvid & ResNet-50 & \checkmark & $67.1$ & $91.8$ \\
DeepLabV3+ & 16 & Camvid & MobileNetV2 & & $65.0$ & $91.7$ \\
DeepLabV3+ & 16 & Camvid & MobileNetV2 & \checkmark & $66.9$ & $91.9$ \\
DeepLabV3+ & 16 & Camvid & ResNet-50 & & $68.2$ & $92.3$ \\
DeepLabV3+ & 16 & Camvid & ResNet-50 & \checkmark & $\mathbf{69.1}$ & $\mathbf{92.3}$ \\
\hline \hline
\multicolumn{7}{c}{Cityscapes validation set}\\
\hline
DeepLabV3+ & 16 & CS-0.2 & MobileNetV2 & & $62.1$ & $95.0$ \\
DeepLabV3+ & 16 & CS-0.2 & MobileNetV2 & \checkmark & $63.8$ & $95.3$
\\
DeepLabV3+ & 16 & CS-0.2 & ResNet50 & & $67.7$ & $95.7$ \\
DeepLabV3+ & 16 & CS-0.2 & ResNet50 & \checkmark & $\mathbf{69.4}$ & $\mathbf{96.0}$
\\ \hline
DeepLabV3+ & 16 & CS-0.5 & MobileNetV2 & & $67.0$ & $95.6$ \\
DeepLabV3+ & 16 & CS-0.5 & MobileNetV2 & \checkmark & $68.1$ & $95.8$
\\
DeepLabV3+ & 16 & CS-0.5 & ResNet50 & & $71.7$ & $96.2$ \\
DeepLabV3+ & 16 & CS-0.5 & ResNet50 & \checkmark & $\mathbf{73.0}$ & $\mathbf{96.3}$
\\
\hline \hline
\end{tabular}
\end{center}
\caption{Overall results on Camvid and Cityscapes (CS) datasets. }
\label{tab:b}
\end{table*}

To understand the advantage of the proposed method, we train FCNs and DeepLab v3+ models using the TAF learning paradigm on partially labelled clips and compare against fully supervised training using only key frames, on both Camvid and Cityscapes dataset. We use mean-average-error (L1 norm) for the contrastive loss as it is a common choice of image applications \footnote{This choice is also discussed in the supplementary material}. We find it important to first normalize the per-pixel prediction via softmax function to avoid learning degenerate features. Our training hyper-parameters are detailed in the supplementary material. We ensure to use a comparable set of hyper-parameters for both the baseline and our method whenever is applicable. For FCN8s, we show results for performing feature swapping only on the stride-32 branch as this leads to better accuracy, while for DeepLab v3+ all branches are swapped with equal probabilities. Our main results are summarized in Table \ref{tab:b}. The main finding is that our method improves over the respective baseline methods using both segmentation algorithms and on both datasets. We note that TAF learning only affects the training time procedures. At inference time models from TAF has exactly the same complexity as the baselines, ensuring that the improvements from TAF is not resultant from increased complexity.




\subsection{Ablation Studies}
\label{sec:alb}

\begin{figure}[hbtp]
    \centering
    \setlength{\abovecaptionskip}{0pt}
    \begin{minipage}[b]{0.49\textwidth}
        \centering
        \includegraphics[width=0.95\textwidth]{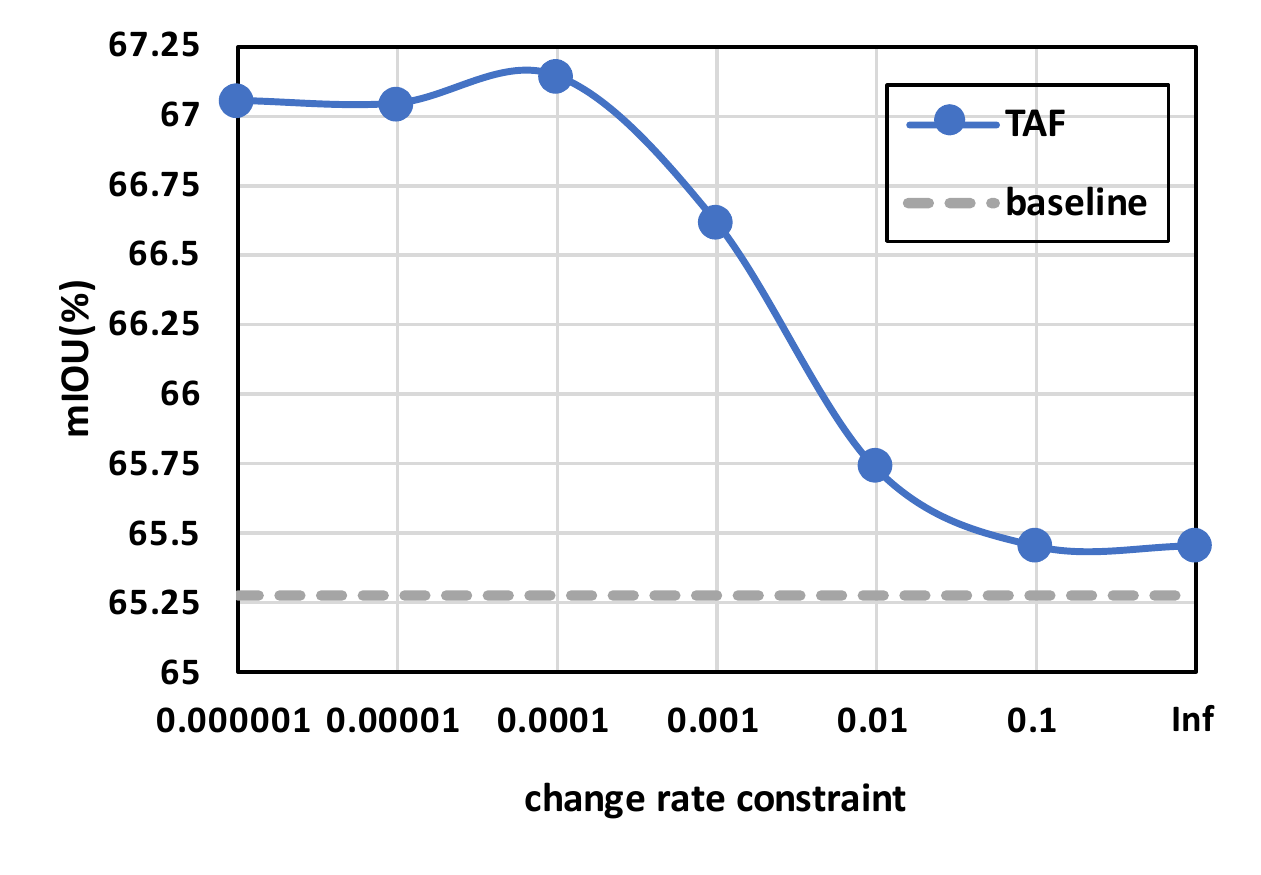}
        \caption{Effect of varying the change rate.}
        \label{fig:varychangerate}
    \end{minipage}
    \begin{minipage}[b]{0.49\textwidth}
        \centering
        \includegraphics[width=0.95\textwidth]{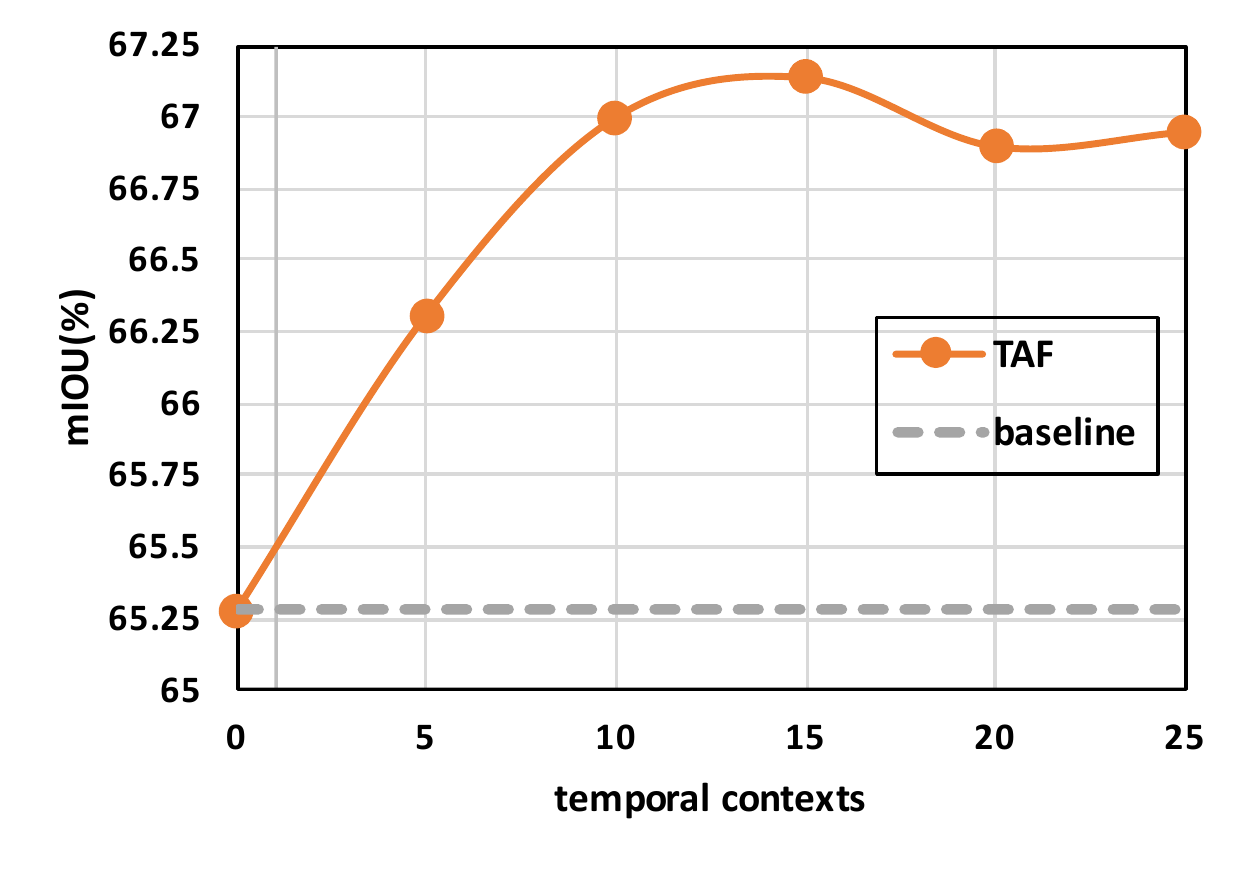}
        \caption{Effect of varying temporal contexts}
        \label{fig:varytemporalcontext}
    \end{minipage}
\end{figure} 

\begin{figure}[hbtp]
    \centering 
    \begin{subfigure}[b]{0.49\textwidth}
        \centering
        \setlength{\abovecaptionskip}{0pt}
        \includegraphics[width=0.95\textwidth]{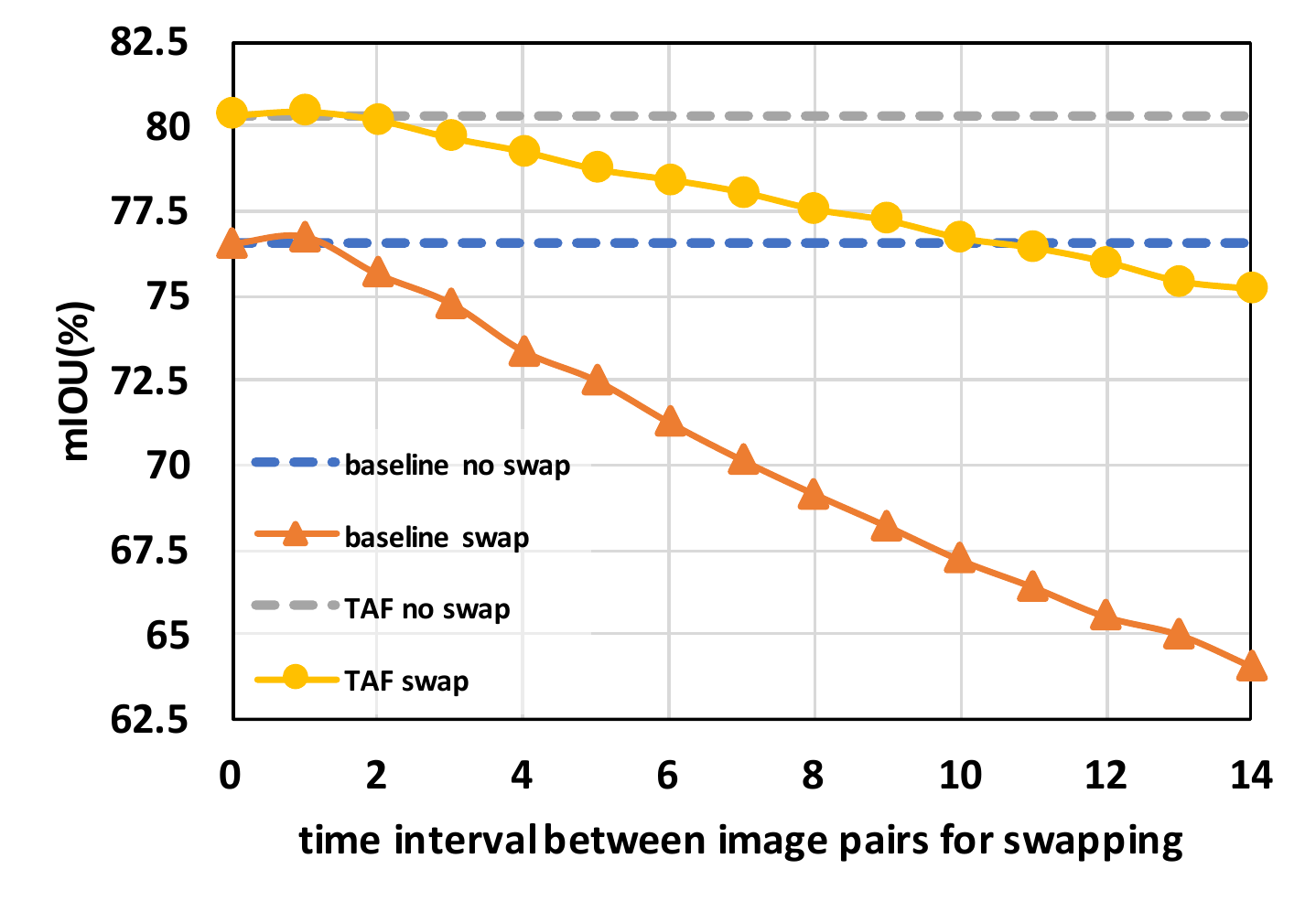}
        \caption{Result on train set}
        \label{fig:swapfeature_train}
    \end{subfigure}
    \begin{subfigure}[b]{0.49\textwidth}
        \centering
        \setlength{\abovecaptionskip}{0pt}
        \includegraphics[width=0.95\textwidth]{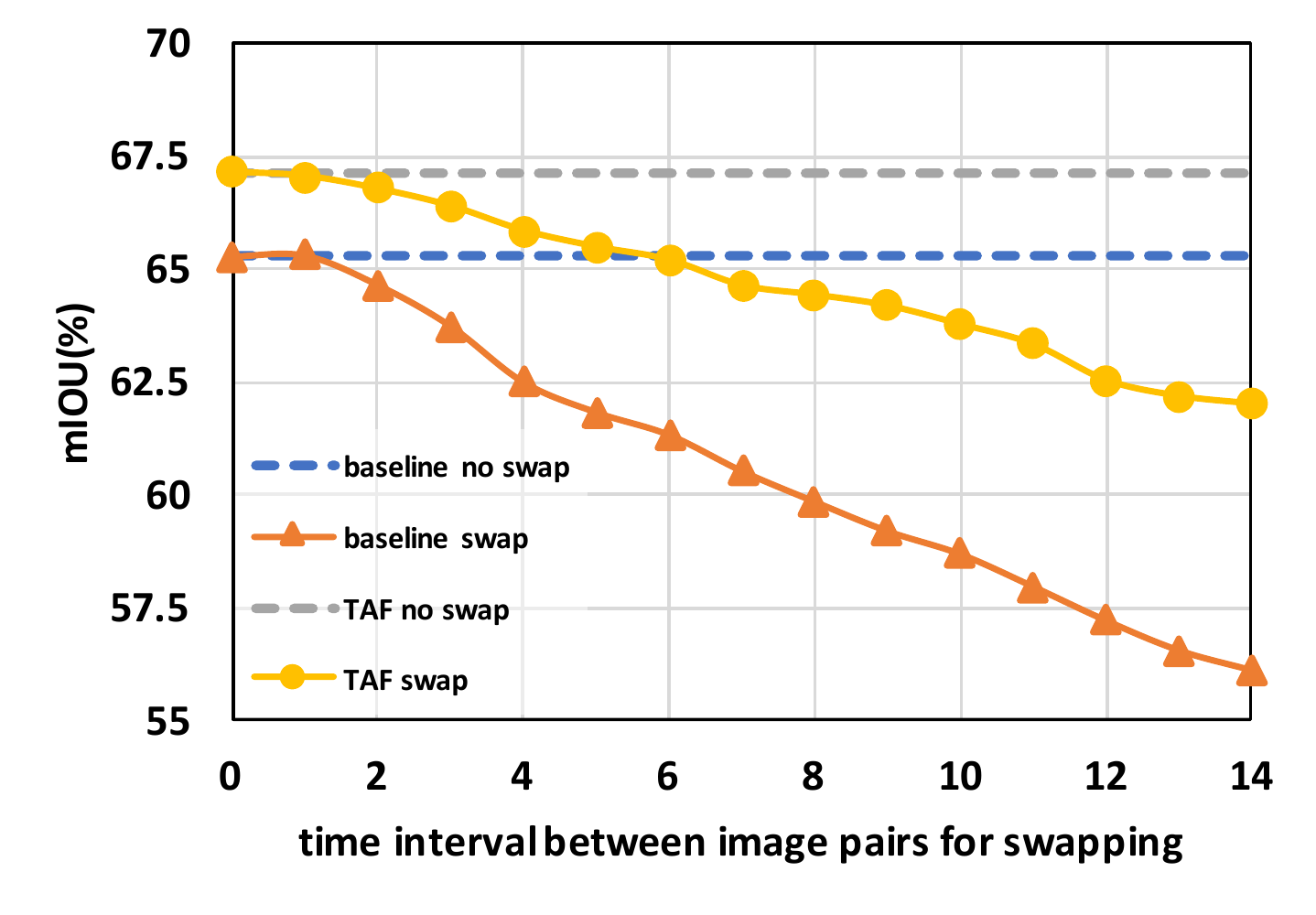}
        \caption{Result on test set}
        \label{fig:swapfeature_test}
    \end{subfigure}
    \caption{Prediction mIOU as a function of feature swapping.}
    \label{fig:swapfeatures}
\end{figure} 

To further understand the proposed method, we perform ablation studies on Camvid dataset using FCN8s models trained with TAF. Feature swapping is only performed on the stride-32 branch for simplicity. We choose to perform ablation studies on this model as its simple design can lead to clearer insights. We use $c_1 = 0.0001$ and the half size of each data clip (a.k.a. length of temporal context) to $15$ unless specified otherwise in particular studies. 

\paragraph{Study on Change Rates Constraint} In our formulation $c_i$ controls change rates of a particular feature dimension. It is interesting to observe the model performance as a function of $c_i$ as this directly informs us on whether the change rate constraint is effective or not. When $c_i$ is $0$, the feature dimension in question will be forced to stay constant across frames. This should force it to learn features that are not informative to the final prediction, effectively reducing the model capacity and consequently, the prediction accuracy. On the other hand, as $c_i$ goes to infinity the regularization term is effectively ignored, leading to sub-optimal results if the proposed regularization is indeed effective.  Our finding as summarized in Figure \ref{fig:varychangerate} is as expected, verifying that the temporal change rate constraints are not trivially imposed. 

\paragraph{Study on Temporal Context} Temporal context refers to how far apart in time a pair of training examples can be. Note that in our derivation, we assume that the pair of frames used in the constraints are sufficiently close. This is important as when the two data points are too far apart, the first order approximation become ineffective. On the other extreme, when setting the temporal context to near zero, the regularization effect is diminished. Results from varying the temporal context are summarized in Figure \ref{fig:varytemporalcontext}. The mIOU reaches maximum when the temporal context is roughly 15 examples (the same setting used in our main results). Interestingly, while a larger temporal context leads to sub-optimal results, the degradation remains relatively mild, suggesting that precise approximation is not critical.

\begin{figure}[hbtp]
    \centering
    \includegraphics[width=\linewidth]{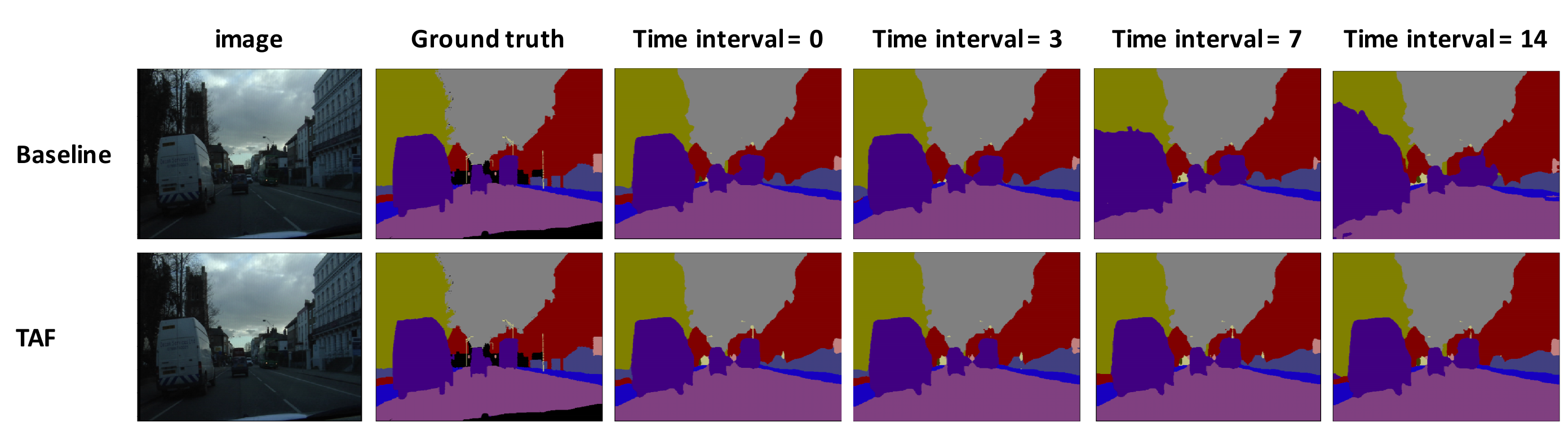}
    \setlength{\abovecaptionskip}{-5pt}
    \caption{Visualization of predictions with feature swapping.}
    \label{fig:swapfeatures_show}
\end{figure}

\paragraph{Study on Feature Swapping} It is particularly interesting to see how the swapping of features between image pairs affects the prediction results. There are two trivial cases that deserve careful consideration: a) If the performance of the model (especially the baseline model) does not show a decrease in accuracy even with feature swapping, then our proposed constraints are not useful as this would suggest a natural tendency for part of the model to learn features that are insensitive to temporal changes. b) If the performance of the model does decrease after feature swapping, but both the baseline model and the TAF models demonstrate similar rate of degradation, then it would cast questions on whether the proposed constraint can actually be successfully imposed in the optimization. Furthermore, whether these constraints imposed on the training set can generalize to a test set. In Figure \ref{fig:swapfeatures} we show that TAF learning does not result in the aforementioned trivial cases and can indeed generalize to held-out sets. Figure \ref{fig:swapfeatures_show} further illustrates the effects of feature swapping. 

\begin{figure}[hbtp]
    \centering
    \includegraphics[width=\linewidth]{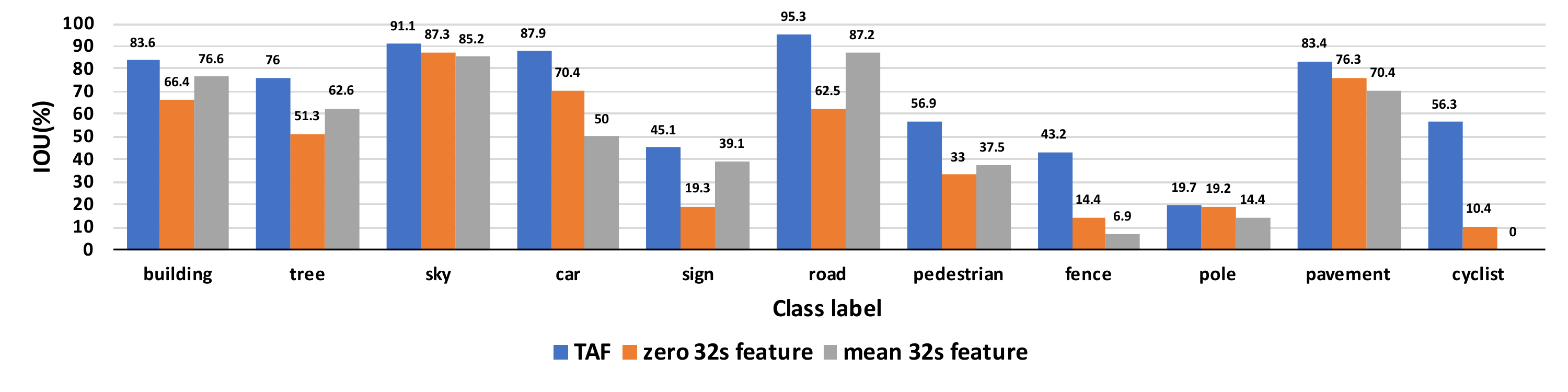}
    \setlength{\abovecaptionskip}{-5pt}
    \caption{Change in prediction IOU after attenuating the stride-32 features. }
    \label{fig:swapfeatures_ablation}
\end{figure}

\paragraph{Study on Feature Attenuation} Constraining the temporal change rate in features could lead to trivial solutions that are not discriminative \cite{7410822}. In Figure \ref{fig:swapfeatures_ablation} we show the effect of replacing the stride-32 branch either with its sample mean or zeros. The large resultant reduction in per-class accuracy suggests that TAF learning is not producing constant, trivial features as feared. However, this issue can be a function of model architectures and should be investigated further in future works. 

\section{Conclusion and Future Works}
\label{sec:conclusion}
In this work, we propose to learn temporally-adaptive features to utilize partially annotated clips. Our proposed framework has demonstrated convincing gains on the challenging task of semantic segmentation. The ablation studies verify that our approach is learning non-trivial features that reflect the proposed temporal rate change constraints, validating our design choices. Our finding suggests the potential of such constraints in enabling self-supervised learning from clip data. It would be interesting to further validate the utility of this approach in related applications. Dense prediction tasks are natural starting points. Another interesting direction is to explore data-driven metrics, such as perceptual loss \cite{perceptual_cvpr18,Johnson2016PerceptualLF,1284395} and adversarial training \cite{NIPS2014_5423}, in constructing the contrastive loss, replacing the current heuristic choice of L1 norm. It is also interesting to further explore existing ideas from slow feature learning and video propagation (explicit label augmentation) to better model problem structures, which may in return lead to stronger results. 

\newpage
{\small
\bibliographystyle{unsrt}
\bibliography{egbib}
}

\newpage
\newpage

\section*{Supplmentary Materials for \ourtitle}

\subsection{Temporal Change Rates of Semantic Classes}

\begin{figure*}[!htbp]
    \centering
    \includegraphics[width=0.9\linewidth]{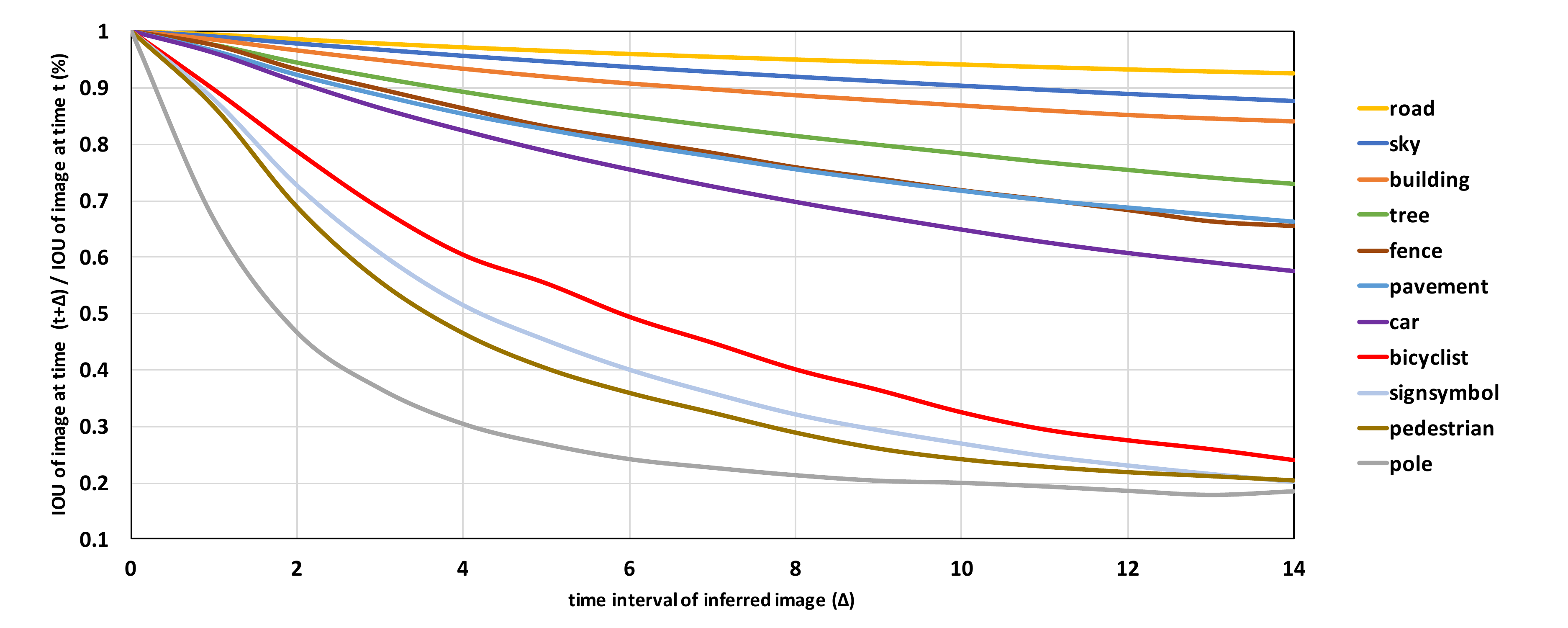}
    \caption{Changing rate of each class in Camvid train set. The figures show the ratio of prediction accuracy on $x_{t}$, between results using features from $x_{t+\Delta}$ and $x_{t}$, measured in per-class IOU. }
    \label{fig:changerateclass}
\end{figure*} 

We expect different semantic classes to demonstrate different temporal change rates. This as we discussed is a motivation for testing our method on semantic segmentation. To verify it empirically, we compare the predicted segmentation labels at $x_{t+\Delta}$ against the ground truth label at $x_t$. Accuracy are reported using IOU normalized by the prediction accuracy at $x_t$, as shown in Figure \ref{fig:changerateclass}. This normalization is necessary as different semantic classes have different intrinsic difficulties. Since labels are not available beyond the key frames, we use the model prediction instead in our study. Interestingly, there is a clear differentiation in the temporal change rates among different classes, as demonstrated by the large differences in change rates of the normalized IOU. Notably, the accuracies of larger or static objects such as road, sky, tree, fence, pavement tend to decrease slowly with time, suggesting low temporal change rates for those structures. On the other hand, smaller or moving objects like car, bicyclist, sign-symbol, pedestrian, pole tend change much faster with time. 

\subsection{Details of Training Procedures}

We use mini-batch SGD optimizer with Nesterov momentum. We set momentum to $0.9$ and weight decay to $0.0001$. The batch size is 16 except for training models with ResNet50 backbone on Cityscapes, in which case due to GPU memory constraints we use batch size of 8. The training are performed on 4 Nvidia GTX 1080Ti GPUs. Synchronized batch normalization \footnote{Implementation: \url{https://github.com/vacancy/Synchronized-BatchNorm-PyTorch}} is used since the number of images per GPU is small in our setting. The ResNet-50 \footnote{Downloaded from \url{https://download.pytorch.org/models/resnet50-19c8e357.pth}} \cite{He_2016_CVPR} and MobileNet v2 \footnote{Downloaded from \url{http://jeff95.me/models/mobilenet_v2-6a65762b.pth}} \cite{Sandler_2018_CVPR} models are pre-trained on ImageNet \cite{ILSVRC15}. We adopt a learning schedule with polynomial decay with power set to $0.9$, following standard practice in semantic segmentation. This schedule multiplies the initial learning rate by the factor $(1 - \frac{current\ epoch}{max\ epoch})^{power}$. During training, we apply random horizontal flip, random scales between $0.5$ and $2$ and random cropping. For both baselines and the TAF models, we report the best results among the initial learning rate from $\{0.005, 0.01, 0.02, 0.025, 0.05\}$ and additionally for ATF learning $\lambda$ from $\{0.5, 1, 2\}$. Additional details are summarized in Table \ref{tab:experiment_details}.

\begin{table}[hbtp]
\begin{center}
\scriptsize 
\setlength{\belowcaptionskip}{-10pt}
\setlength\tabcolsep{3pt}
\renewcommand{\arraystretch}{1.25}
\begin{tabular}{lccccccccc}
\hline \hline
\textbf{Method} & \textbf{Training set} & \textbf{Backbone} & \textbf{TAF} & \textbf{Init. lr} & $\mathbf{\lambda}$ & \textbf{Init. Img size} & \textbf{Crop size} & \textbf{Epoch} & \textbf{Val size} \\ 
\hline 
FCN8s & Camvid & ResNet-50 & & 0.02 & N/A & $360 \times 480$ & $360 \times 360$ & 600 & $360 \times 480$ \\
FCN8s & Camvid & ResNet-50 & \checkmark & 0.02 & 1.0 & $360 \times 480$ & $360 \times 360$ & 600 & $360 \times 480$ \\
DeepLabV3+ & Camvid & MobileNetV2 &  & 0.02 & N/A & $360 \times 480$ & $360 \times 360$ & 600 & $360 \times 480$ \\
DeepLabV3+ & Camvid & MobileNetV2 & \checkmark & 0.05 & 1.0 & $360 \times 480$ & $360 \times 360$ & 600 & $360 \times 480$ \\
DeepLabV3+ & Camvid & ResNet-50 & & 0.02 & N/A & $360 \times 480$ & $360 \times 360$ & 600 & $360 \times 480$ \\
DeepLabV3+ & Camvid & ResNet-50 & \checkmark & 0.05 & 0.5 & $360 \times 480$ & $360 \times 360$ & 600 & $360 \times 480$ \\
\hline 
DeepLabV3+ & CS-0.2 & MobileNetV2 & & 0.02 & N/A & $1024 \times 2048$ & $720 \times 720$ & 300 & $1024 \times 2048$ \\
DeepLabV3+ & CS-0.2 & MobileNetV2 & \checkmark & 0.05 & 1.0 & $1024 \times 2048$ & $720 \times 720$ & 300 & $1024 \times 2048$ \\
DeepLabV3+ & CS-0.2 & ResNet50 & & 0.02 & N/A & $1024 \times 2048$ & $720 \times 720$ & 300 & $1024 \times 2048$ \\
DeepLabV3+ & CS-0.2 & ResNet50 & \checkmark & 0.05 & 1.0 & $1024 \times 2048$ & $720 \times 720$ & 300 & $1024 \times 2048$ \\
\hline
DeepLabV3+ & CS-0.5 & MobileNetV2 & & 0.02 & N/A & $1024 \times 2048$ & $720 \times 720$ & 300 & $1024 \times 2048$ \\
DeepLabV3+ & CS-0.5 & MobileNetV2 & \checkmark & 0.02 & 1.0 & $1024 \times 2048$ & $720 \times 720$ & 300 & $1024 \times 2048$ \\
DeepLabV3+ & CS-0.5 & ResNet50 & & 0.01 & N/A & $1024 \times 2048$ & $720 \times 720$ & 300 & $1024 \times 2048$ \\
DeepLabV3+ & CS-0.5 & ResNet50 & \checkmark & 0.01 & 1.0 & $1024 \times 2048$ & $720 \times 720$ & 300 & $1024 \times 2048$ \\
\hline \hline
\end{tabular}
\end{center}
\setlength{\belowcaptionskip}{-10pt}
\caption{Hyperparameters used for main results.  }
\label{tab:experiment_details}
\end{table}



\subsection{Choice of Temporal Change Rates}
For FCN8s, our preliminary studies suggest that assigning $c_1 = 0.0001$ and $c_2, c_3$ to $\inf$ leads to best performance. We note that this design effectively disables TAF learning on the stride-16 and stride-8 branches. This design is necessary to allow the two high resolution features to model structures with fast temporal change rates sufficiently. For DeepLab v3+, we assign $c_1=0.000001, c_2=0.000001, c_3=0.0001, c_4=0.001, c_5=0.01$. There is no advantage in disabling TAF learning on any branch. In fact, our study suggests that it leads to worse accuracy. We think that can be attributable to the decoder ($\Omega$ function) design of the DeepLab v3+, which provides a skip connection with output stride of $4$ from low level features and can model fast features sufficiently by itself. 

\subsection{Measure Temporal Change Rates Relative to Input}
In our formulation the temporal change rates are directly measured by the variations in the predictive model $y$. However, different clips can have intrinsically different rates of motions, thus it might be wise to impose the temporal change rate constraints relative to the change rates in input images. This, as we also discuss in Section \ref{sec:method}, is not trivial since $\frac{dx}{dt}$ is not available. In our preliminary studies, we empirically test using $L1$ norm as a measure of the change rate, using the first order approximation as we do for $y$. Then, we set the constraints as the proportion between the change rates in $y$ and those in $x$. We find that this does not lead to improvement over the design we presented and the training is usually less stable. We conjecture that this is attributable to our heuristic method in measuring differences between images, a point worth revisiting in future works. 

\subsection{Choice of Loss Functions}

The loss function consists of two parts: The semantic loss function and the contrastive loss. The former compares the prediction of the model against the ground truth annotations at the key frames, while the latter compares the prediction from the model before and after feature swapping (our regularization term). For the semantic loss function, we use cross-entropy loss for both the baseline and TAF learning, per standard practice. For the contrastive loss, in our preliminary studies we experiment with a few different metrics, including mean-squared-error (MSE), mean-average-error (MAE, or L1 loss) as well as symmetric cross entropy. Among them, L1 loss leads to more stable training and best results. We note that it is important to first normalize the prediction at every pixel via a softmax function, as applying L1 norm regularization directly on the logits (before normalization) leads to degenerate solutions. We note that L1 norm is by no means the optimal choice of the metric to compare images, as it does not reflect the rich semantic structures encoded in natural images. We believe that perceptual loss or even adversarial loss (via a learnable model) could lead to better performance and are interesting future directions to explore. 

\end{document}